\documentclass[keynote]{sigirforum}

\def\pubissue{Vol. 61 No. 1 -- December 2025}

\def\keynotedate{8 April 2025}

\begin{document}
\title{Understanding the Interplay between LLMs' Utilisation of Parametric and Contextual Knowledge: A keynote at ECIR 2025}

\authors{
\author[augenstein@di.ku.dk]{Isabelle Augenstein}{University of Copenhagen}{Denmark}
}

\maketitle 
\begin{abstract}
Language Models (LMs) acquire parametric knowledge from their training process, embedding it within their weights. The increasing scalability of LMs, however, poses significant challenges for understanding a model's inner workings and further for updating or correcting this embedded knowledge without the significant cost of retraining. Moreover, when using these language models for knowledge-intensive language understanding tasks, LMs have to integrate relevant context, mitigating their inherent weaknesses, such as incomplete or outdated knowledge. Nevertheless, studies indicate that LMs often ignore the provided context as it can be in conflict with the pre-existing LM's memory learned during pre-training. Conflicting knowledge can also already be present in the LM's parameters, termed intra-memory conflict. This underscores the importance of understanding the interplay between how a language model uses its parametric knowledge and the retrieved contextual knowledge.
In this talk, I will aim to shed light on this important issue by presenting our research on evaluating the knowledge present in LMs, diagnostic tests that can reveal knowledge conflicts, as well as on understanding the characteristics of successfully used contextual knowledge.
\end{abstract}

\section{Introduction}
LLM usage has become ubiquitous, with people using LLMs for a wide range of both creative and information-seeking tasks \citep{conf/iclr/Zhao0HC0D24,CHIARELLO2024103002}. 
In some domains such as research, information seeking is the key usage, with LLMs being used for discovering papers, generating summaries or explanations, asking broad questions about a field, or discovering topics \citep{journals/corr/abs-2411-05025}. 

Given the wide-ranging capabilities of LLMs, there have been suggestions of LLMs displaying Artificial General Intelligence (AGI), meaning the possession of human-like cognitive abilities. However, upon closer investigation, it becomes clear that while LLMs show high performance generally, they display several fundamental shortcomings \citep{bang-etal-2023-multitask}. High amounts of dataset contamination \citep{xu-etal-2025-infini} mean that LLMs have been trained on the benchmark datasets they are evaluated on already, thus not having to perform reasoning, but merely the recall of existing answers. Thus, drastic performance drops can be observed when performing small alterations to the wording used in those benchmark datasets \citep{journals/corr/abs-2504-00509,mizrahi-etal-2024-state}. LLMs have also shown to perform poorly on low-resource languages \citep{pava2025mind}, at most types of reasoning \citep{shojaee2025illusionthinkingunderstandingstrengths}, and display many factual errors due to a lack of access to a knowledge base \citep{journals/natmi/AugensteinBCCCCDFHHHJMM24}. 
These issues partly arise because LLMs are developed as general-purpose models for both creative and information-seeking tasks -- for the former, hallucinations might even be desirable, whereas for the latter, they are to be avoided at all costs.

There are a number of avenues that have been explored to improve the factuality of language models. Their consistency can be improved using methods including chain-of-thought prompting \citep{wei2022chainofthought}, self-consistency checking \citep{conf/iclr/0002WSLCNCZ23}, continual learning \citep{10.1145/3735633}, and knowledge editing \citep{10.1145/3698590}. 
However, these methods all have inherent downsides. Self-consistency checking and knowledge editing is challenging as models are inherently not very consistent due to issues such as the above-mentioned prompt instability. Continual learning is by nature highly costly.
Knowledge editing can lead to ripple effects, i.e. the editing of knowledge beyond what is intended \citep{cohen-etal-2024-evaluating} as well as the removal of long-tail knowledge, due to knowledge superposition \citep{10.1609/aaai.v39i22.34583}, and identifying what knowledge to edit is a task in and of itself \citep{journals/corr/abs-2508-08879}. Thus, internal consistency checking only partly addresses the factuality issues of LLMs.
Another direction is the combination with external knowledge, by detecting and correcting factual mistakes at inference time \citep{wang-etal-2024-factcheck}, using a modularised knowledge-grounded framework \citep{arakelyan-etal-2025-flare}, or, very commonly, retrieval-augmented generation (RAG) \citep{10.1145/3637528.3671470}. These can better take the context-dependent nature of queries into account, by retrieving \textit{contextual knowledge} to augment the LLM’s \textit{parametric knowledge}. 

However, a key research gap is that the interplay between contextual and parametric knowledge underexplored, as is when contextual knowledge even should overwrite parametric knowledge. This keynote explores these two topics by highlighting recent research conducted in our research group.

\section{Determining what Parametric Knowledge influences a LLM’s Prediction}

\begin{figure}[!t]
\begin{center}
\includegraphics[width=0.65\columnwidth]{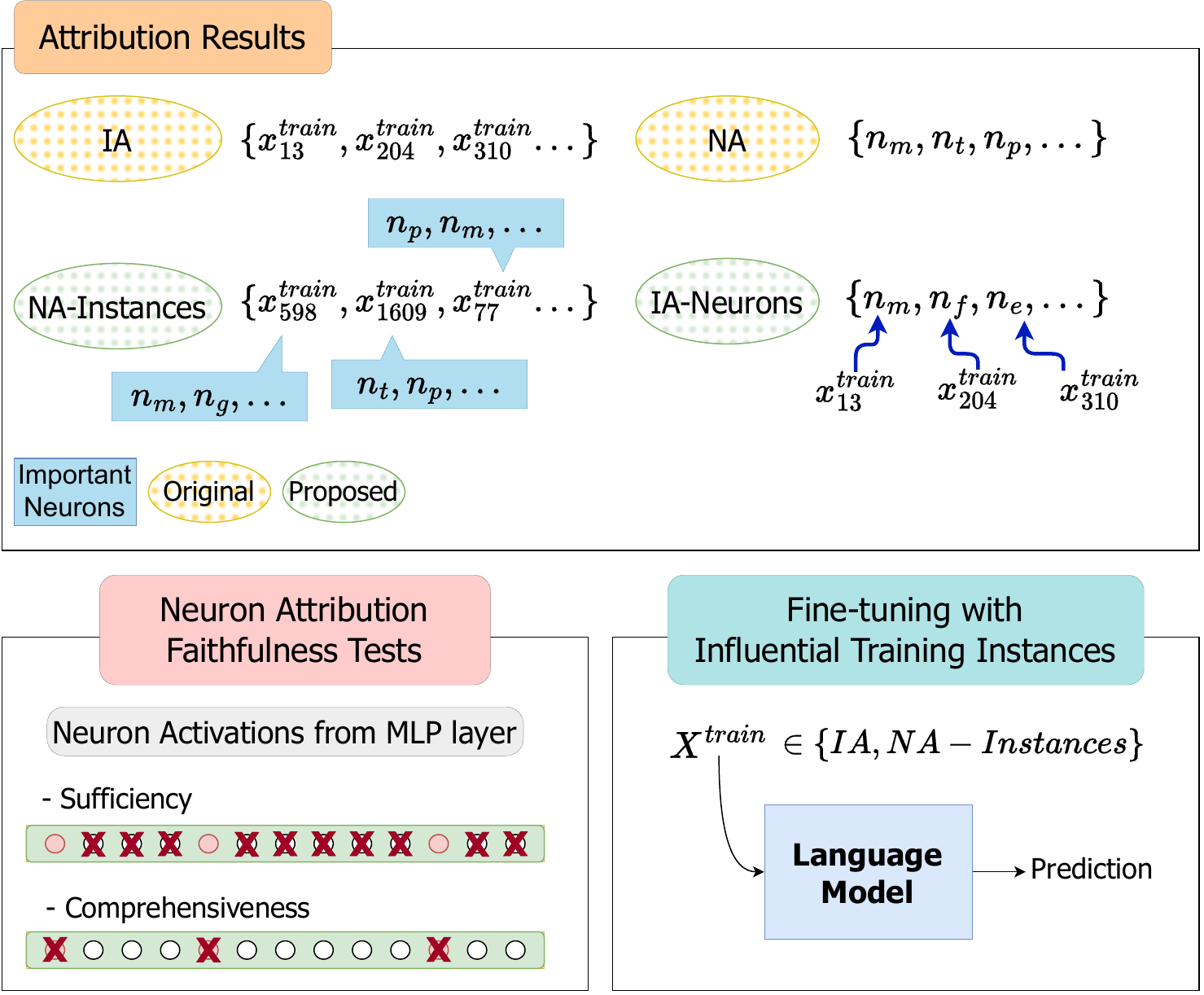}
\caption{Evaluation framework comparing Instance and Neuron Attribution methods \citep{yu-etal-2024-revealing}.} 
\label{fig1}
\end{center}
\end{figure}

\textit{Parametric knowledge} is the knowledge encoded in a LM's weights which an LM acquires during training.  
Our study \citep{yu-etal-2024-revealing} focuses on the changes in knowledge acquired during LLM training and task-adaptive training for knowledge-intensive tasks, such as fact checking, QA, and natural language inference. 
To unveil LM’s parametric knowledge used to arrive at a prediction, attribution methods are commonly used. Previous methods operate on different levels (e.g. instance vs. neuron) and are studied in isolation, with no consensus as to which methods work best best in which scenarios. We thus propose a unified evaluation framework, illustrated in Figure \ref{fig1}, that compares these two streams of attribution methods, to provide a comprehensive understanding of a LM's inner workings. The framework includes methods to align the most influential training instances with the most important neurons. The success of this alignment procedure is evaluated using faithfulness tests for sufficiency (i.e. the activation of key neurons) and completeness (the suppression of the activation of key neurons), along with an evaluation of the impact of fine-tuning with influential training instances.

We analyse the MLP classification layers instead of entire model weights due to the high computational cost, as prior worked shows that this layers shows a high correlation with the overall model weights \citep{pezeshkpour-etal-2021-empirical}.
Our findings show that most MLP neurons can be removed without significant changes to the predictions compared to the original model, contradicting the assumption of prior work that most model's knowledge is located in these neurons. We hypothesise that this might be due to the importance of attention weights for encoding knowledge, as also argued by prior work \citep{wiegreffe-pinter-2019-attention}.
Our experiments on fine-tuning with influential training instances identified using Instance Attribution leads to results on par with the equivalent number of randomly selected training instances. A regression test reveals that this is likely due to the diversity of highly influential training instances being significantly lower than that of randomly selected training instances, which hampers the effectiveness in reaching a high performance with the fine-tuned model.
However, we find a potential application in combining Instance and Neuron Attribution for the discovery of dataset artifacts \citep{mccoy-etal-2019-right}, specifically models overfitting to certain lexical patterns observed at training time, with the combined methods being more effective at discovering such instances than either Neuron or Instance Attribution alone.
Overall, we find that Instance Attribution and Neuron Attribution result in different explanations about the knowledge responsible for the test prediction.

\section{Revealing Conflicts between Parametric and Contextual Knowledge}

\begin{figure*}[!t]
    \centering
    \centerline{\includegraphics[width=1\textwidth]{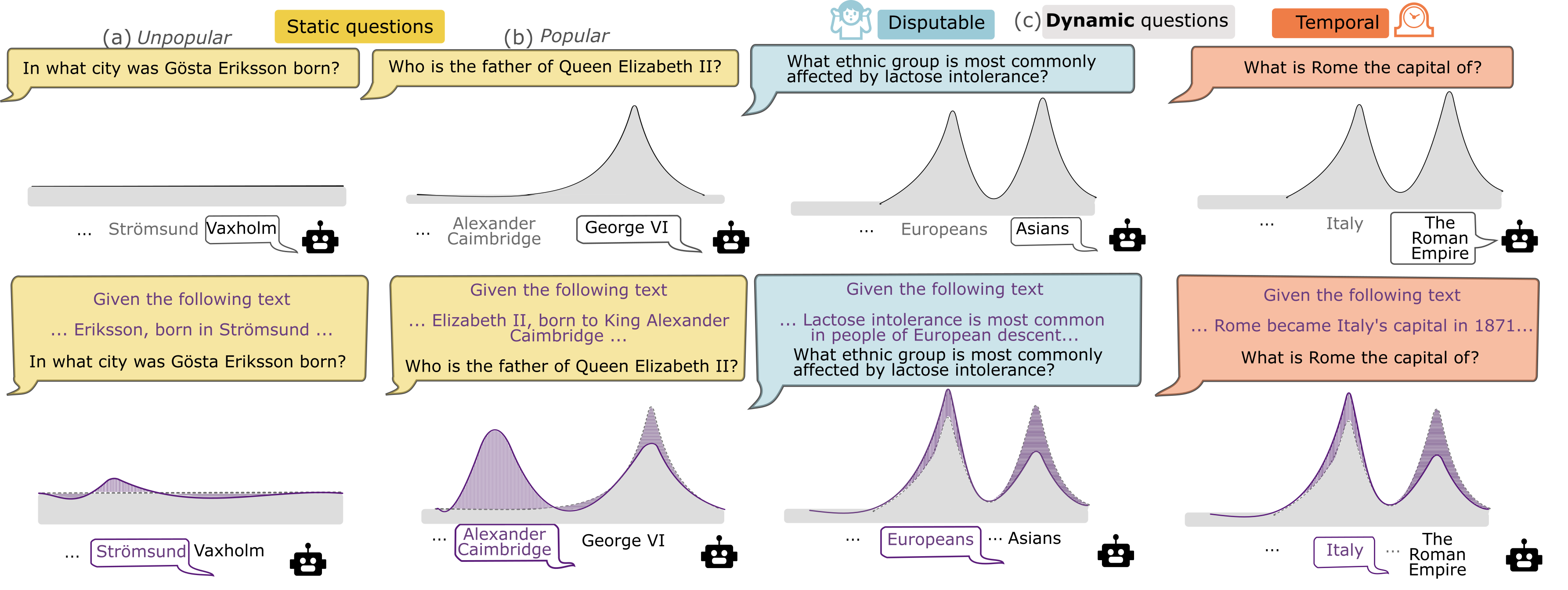}}
    \caption{Instances with varying dynamicity in the \textsc{DynamicQA} dataset \citep{marjanovic-etal-2024-dynamicqa}.} 
    \label{fig:triplets}
\end{figure*}

Though attribution methods methods, as studied in the previous section, can theoretically pinpoint the knowledge a prediction is based on, their application is expensive, only possible given access to full model parameters, and their application to only partial model parameters can result in unintuitive or contradictory findings. An alternative approach is to use probing tests, which we use to study two different types of knowledge conflicts: \textit{intra-memory conflict}, the conflict caused by contradicting representations of the fact within the training data, can cause uncertainty and instability of an LM, and \textit{context-memory conflict}, the conflict caused by the context contradicts to the parametric knowledge \citep{marjanovic-etal-2024-dynamicqa}. We study these two phenomena in the context of question answering, where fact dynamicity can impact the knowledge that should be used. To this end, our dataset \textsc{DynamicQA} contains samples of \textit{static facts}, which only have one possible representation; \textit{temporal facts}, which change over time; and \textit{disputable facts}, which can change depending on the viewpoint (see Figure \ref{fig:triplets}). 

We then study how the output distribution varies depending on these types of facts, as well as when additional context is provided, as is done in RAG settings. Dynamic facts (both temporal and disputable ones) should then lead to intra-memory conflicts, and presenting models with conflicting context during RAG should lead to context-memory conflicts. We also expect to observe differences between popular facts, for which a model is expected to be much more certain than unpopular facts. 
We measure intra-memory conflict using semantic entropy, which captures the semantic variation present in parametric memory \citep{kuhn2023semantic}, and context-memory conflict using a measure we term \textit{coherent persuasion score}, based on \citet{du2024context} to approximate an LM's semantic shift in output distribution given competing context.

Our findings reveal that, counter-intuitively, presenting LMs with manipulated contexts of static facts and facts with low dynamicity results in greater persuasiveness for LMs than dynamic facts -- or, put differently, facts that change regularly are less likely to be updated with context-retrieval, yet facts that never change are easily persuaded. This fact dynamicity is found to be the strongest, most consistent negative indicator of model persuasion across models, outperforming fact popularity, which was previous used to guide RAG models. These results underline the need for new measures of intra-memory conflict and the need for other indicators of successful context usage in RAG, especially for low-certainty domains \citep{ni2024llmsneedretrievalaugmentation}.

\section{Determining when or how RAG uses Contextual Knowledge}

\begin{figure}[!t]
    \centering
    \includegraphics[width=0.65\columnwidth,trim={1.9cm 3.9cm 25.3cm 0.3cm},clip]{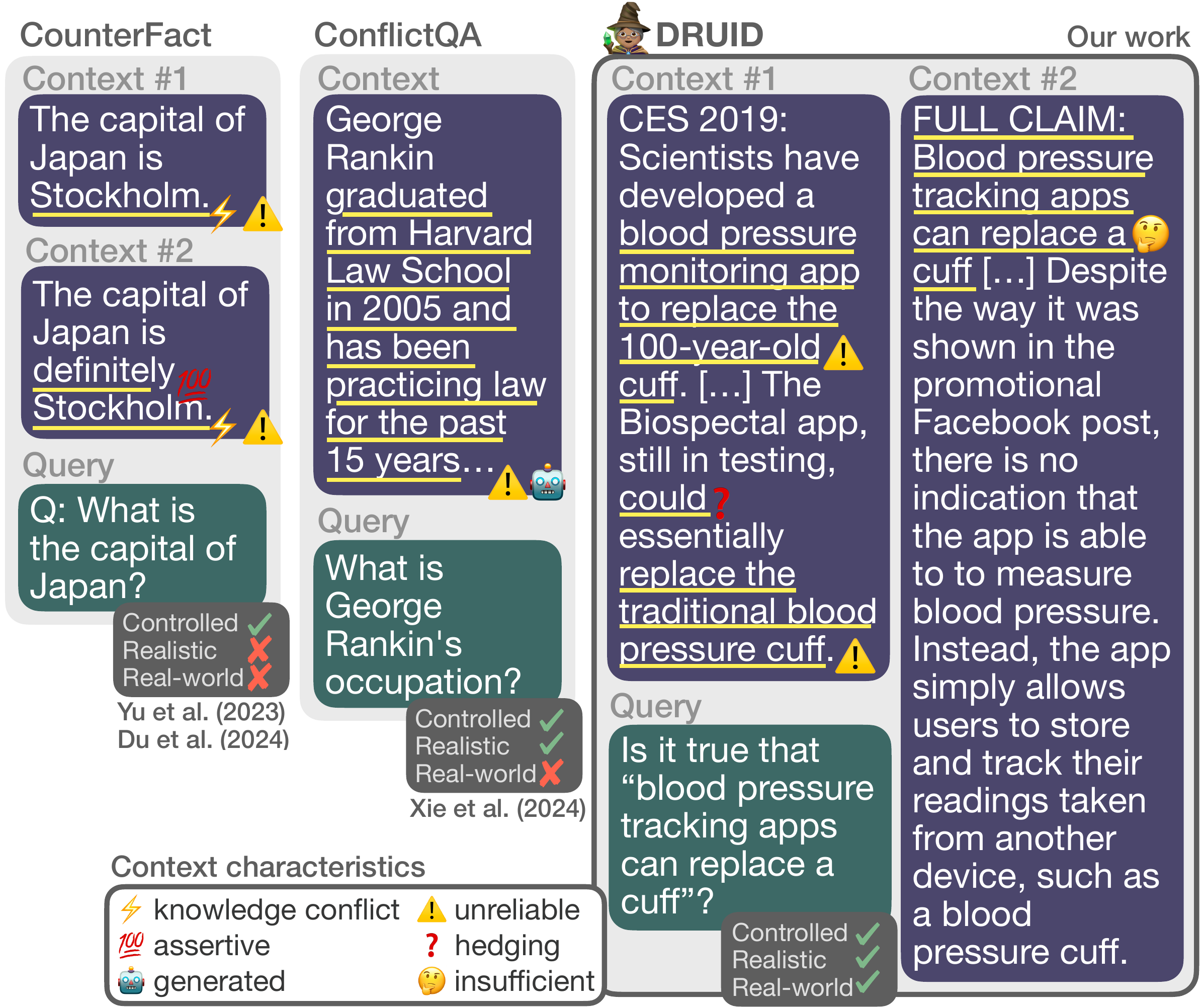}
    \caption{The \textsc{DRUID} dataset \citep{hagstrom-etal-2025-reality} in comparison with prior synthetic datasets.}
    \label{fig:comparison-overview}
\end{figure}

Successful RAG not only relies on the usage of retrieved information, but also on the retrieval of relevant information, and the interplay between these two components, though prior work studies these aspects in isolation. 
As such, little is understood about the characteristics of retrieved content, and its impact on LLM usage. Though some context usage studies exist, they use synthetic data, and thus do not reflect real-world RAG scenarios.
To overcome this, we propose new dataset for claim verification with real-world contexts to measure realistic context usage (\textsc{DRUID}), a novel context usage measure (ACU), and generate novel insights into LLMs’ context usage characteristics \citep{hagstrom-etal-2025-reality}.

Some patterns we find are that context from fact-check sources leads to high context usage, which is likely due to the higher rate of assertive and to-the-point language. Also, the more direct discussion of claims with multiple arguments might make it more more convincing to the LM. Similarly, evidence documents published after the claim and gold evidence sources lead to higher context usage. Conversely, references to external sources show low correlations with ACU. Moreover, LLMs prioritise contexts with high query-context similarity, which are more difficult to obtain in real-world RAG setting. Lastly, LMs are shown to be less faithful to long contexts.

Comparing \textsc{DRUID} to synthetic datasets, we find that synthetic datasets oversell the impact of certain context characteristics (e.g. knowledge conflicts), which are rare in retrieved data. Synthetic data exaggerates context repulsion, which is rarer for realistic data. While we identify different characteristics indicating RAG failure in real-world settings, there is no single indicative characteristic. This provides a reality check on LLM context usage, and underscores the need for real-world aligned studies to understand and improve context use for RAG. 

In some follow-up work, we study richer forms of interaction between parametric and contextual knowledge for RAG, namely complementary or supporting knowledge \citep{islam2025multistepknowledgeinteractionanalysis}. There, we find that for claim verification, knowledge conflicts often pushes models to prefer contextual rather than parametric knowledge, whereas for common-sense question answering with fewer knowledge conflicts, LLMs more often rely on their parametric memory.

Moreover, we perform a study on how context-memory conflicts can be resolved using a wide range of so-called \textit{context manipulation techniques (CMT)} \citep{hagström2025cubbenchmarkingcontextutilisation}, going beyond simply providing the context as done in \citet{marjanovic-etal-2024-dynamicqa,hagstrom-etal-2025-reality}. These range from simple methods such as prompting to more elaborate methods including mechanistic interventions. The study is performed on the datasets as \citet{hagstrom-etal-2025-reality}. We find that, overall, there is no clear winner between the different CMTs. Moreover, larger LLMs on average perform better than smaller LLMs at successfully using context, though smaller models can outperform them with the best CMT for the specific model.

\section{Concluding Remarks}

Despite ongoing work, the inner workings of RAG-based language models, and how and when they use or ignore context is still poorly understood. This is a unique challenge necessitating collaborations between researchers working on information retrieval and natural language processing. 
This keynote has aimed to shed light on some aspects of this topic, though often, and notwithstanding rigorous evaluation efforts, our research has unearthed more questions than it has provided answers. 

Some effects we observed have been reminiscent of trends observed for earlier neural methods used in NLP, namely pre-trained language models. These include that LLMs are excellent at recitation, not at reasoning \citep{journals/corr/abs-2504-00509}, which could also be observed for PLMs \citep{petroni-etal-2019-language}. The fact that RAG-based claim verification models prioritise easy-to-understand sources \citep{hagstrom-etal-2025-reality} can also be traced back to effects observed for PLMs \citep{augenstein-etal-2019-multifc}.

As this ECIR 2025 keynote was a Karen Spärk Jones award lecture, I would like conclude by honouring her memory in sharing one of her famous quotes: 

\textit{Those […] who had been around for a long time, can see old ideas reappearing in new guises […]. But the new costumes are better made, of better materials, as well as more becoming: so research is not so much going round in circles as ascending a spiral (Karen Spärk Jones, 1994).}

\section*{Acknowledgements}
$\begin{array}{l}\includegraphics[width=1cm]{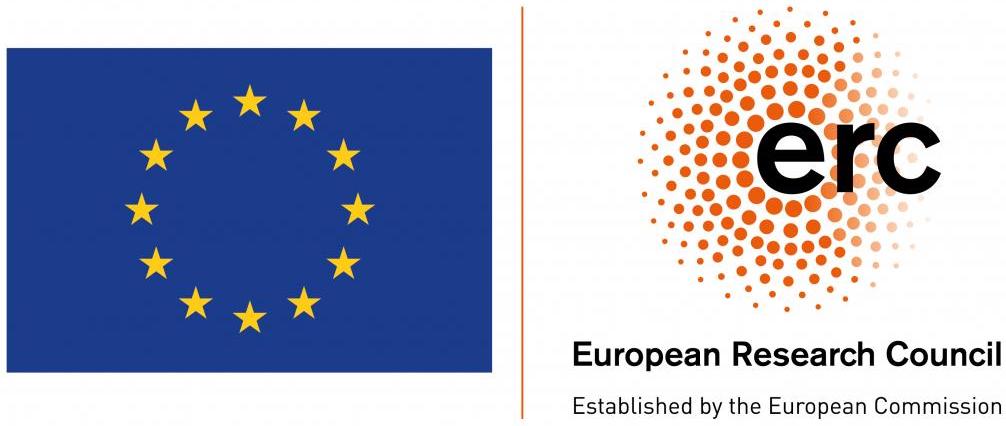} \end{array}$ 
I would like to thank The Chartered Institute for IT (BCS) as well as Bloomberg for honouring me with the Karen Spärck Jones Award, as well as my research group, without whom none of this would have been possible.

This research was in large parts supported by the European Union (ERC, ExplainYourself, 101077481). Views and opinions expressed are however those of the author(s) only and do not necessarily reflect those of the European Union or the European Research Council. Neither the European Union nor the granting authority can be held responsible for them.

\bibliography{sigirforum}

\begin{thebibliography}{33}
\providecommand{\natexlab}[1]{#1}
\providecommand{\url}[1]{\texttt{#1}}
\expandafter\ifx\csname urlstyle\endcsname\relax
  \providecommand{\doi}[1]{doi: #1}\else
  \providecommand{\doi}{doi: \begingroup \urlstyle{rm}\Url}\fi

\bibitem[Arakelyan et~al.(2025)Arakelyan, Minervini, Lewis, Verga, and
  Augenstein]{arakelyan-etal-2025-flare}
Erik Arakelyan, Pasquale Minervini, Patrick Lewis, Pat Verga, and Isabelle
  Augenstein.
\newblock {{FLARE}: Faithful Logic-Aided Reasoning and Exploration}.
\newblock In Christos Christodoulopoulos, Tanmoy Chakraborty, Carolyn Rose, and
  Violet Peng, editors, \emph{Proceedings of the 2025 Conference on Empirical
  Methods in Natural Language Processing}, pages 23396--23414, Suzhou, China,
  November 2025. Association for Computational Linguistics.
\newblock ISBN 979-8-89176-332-6.
\newblock \doi{10.18653/v1/2025.emnlp-main.1193}.
\newblock URL \url{https://aclanthology.org/2025.emnlp-main.1193/}.

\bibitem[Augenstein et~al.(2019)Augenstein, Lioma, Wang, Chaves~Lima, Hansen,
  Hansen, and Simonsen]{augenstein-etal-2019-multifc}
Isabelle Augenstein, Christina Lioma, Dongsheng Wang, Lucas Chaves~Lima, Casper
  Hansen, Christian Hansen, and Jakob~Grue Simonsen.
\newblock {{M}ulti{FC}: A Real-World Multi-Domain Dataset for Evidence-Based
  Fact Checking of Claims}.
\newblock In Kentaro Inui, Jing Jiang, Vincent Ng, and Xiaojun Wan, editors,
  \emph{Proceedings of the 2019 Conference on Empirical Methods in Natural
  Language Processing and the 9th International Joint Conference on Natural
  Language Processing (EMNLP-IJCNLP)}, pages 4685--4697, Hong Kong, China,
  November 2019. Association for Computational Linguistics.
\newblock \doi{10.18653/v1/D19-1475}.
\newblock URL \url{https://aclanthology.org/D19-1475/}.

\bibitem[Augenstein et~al.(2024)Augenstein, Baldwin, Cha, Chakraborty,
  Ciampaglia, Corney, DiResta, Ferrara, Hale, Halevy, Hovy, Ji, Menczer,
  Míguez, Nakov, Scheufele, Sharma, and
  Zagni]{journals/natmi/AugensteinBCCCCDFHHHJMM24}
Isabelle Augenstein, Timothy Baldwin, Meeyoung Cha, Tanmoy Chakraborty,
  Giovanni~Luca Ciampaglia, David P.~A. Corney, Renee DiResta, Emilio Ferrara,
  Scott Hale, Alon~Y. Halevy, Eduard~H. Hovy, Heng Ji, Filippo Menczer, Rubén
  Míguez, Preslav Nakov, Dietram Scheufele, Shivam Sharma, and Giovanni Zagni.
\newblock Factuality challenges in the era of large language models and
  opportunities for fact-checking.
\newblock \emph{Nat. Mac. Intell.}, 6\penalty0 (8):\penalty0 852--863, 2024.
\newblock URL \url{https://www.nature.com/articles/s42256-024-00881-z}.

\bibitem[Bang et~al.(2023)Bang, Cahyawijaya, Lee, Dai, Su, Wilie, Lovenia, Ji,
  Yu, Chung, Do, Xu, and Fung]{bang-etal-2023-multitask}
Yejin Bang, Samuel Cahyawijaya, Nayeon Lee, Wenliang Dai, Dan Su, Bryan Wilie,
  Holy Lovenia, Ziwei Ji, Tiezheng Yu, Willy Chung, Quyet~V. Do, Yan Xu, and
  Pascale Fung.
\newblock {A Multitask, Multilingual, Multimodal Evaluation of {C}hat{GPT} on
  Reasoning, Hallucination, and Interactivity}.
\newblock In Jong~C. Park, Yuki Arase, Baotian Hu, Wei Lu, Derry Wijaya, Ayu
  Purwarianti, and Adila~Alfa Krisnadhi, editors, \emph{{Proceedings of the
  13th International Joint Conference on Natural Language Processing and the
  3rd Conference of the Asia-Pacific Chapter of the Association for
  Computational Linguistics (Volume 1: Long Papers)}}, pages 675--718, Nusa
  Dua, Bali, November 2023. Association for Computational Linguistics.
\newblock \doi{10.18653/v1/2023.ijcnlp-main.45}.
\newblock URL \url{https://aclanthology.org/2023.ijcnlp-main.45/}.

\bibitem[Chiarello et~al.(2024)Chiarello, Giordano, Spada, Barandoni, and
  Fantoni]{CHIARELLO2024103002}
Filippo Chiarello, Vito Giordano, Irene Spada, Simone Barandoni, and Gualtiero
  Fantoni.
\newblock {Future applications of generative large language models: A
  data-driven case study on ChatGPT}.
\newblock \emph{Technovation}, 133:\penalty0 103002, 2024.
\newblock ISSN 0166-4972.
\newblock \doi{https://doi.org/10.1016/j.technovation.2024.103002}.
\newblock URL
  \url{https://www.sciencedirect.com/science/article/pii/S016649722400052X}.

\bibitem[Cohen et~al.(2024)Cohen, Biran, Yoran, Globerson, and
  Geva]{cohen-etal-2024-evaluating}
Roi Cohen, Eden Biran, Ori Yoran, Amir Globerson, and Mor Geva.
\newblock {Evaluating the Ripple Effects of Knowledge Editing in Language
  Models}.
\newblock \emph{Transactions of the Association for Computational Linguistics},
  12:\penalty0 283--298, 2024.
\newblock \doi{10.1162/tacl_a_00644}.
\newblock URL \url{https://aclanthology.org/2024.tacl-1.16/}.

\bibitem[Du et~al.(2024)Du, Sn{\ae}bjarnarson, Stoehr, White, Schein, and
  Cotterell]{du2024context}
Kevin Du, V{\'e}steinn Sn{\ae}bjarnarson, Niklas Stoehr, Jennifer White, Aaron
  Schein, and Ryan Cotterell.
\newblock {Context versus Prior Knowledge in Language Models}.
\newblock In Lun-Wei Ku, Andre Martins, and Vivek Srikumar, editors,
  \emph{Proceedings of the 62nd Annual Meeting of the Association for
  Computational Linguistics (Volume 1: Long Papers)}, pages 13211--13235,
  Bangkok, Thailand, August 2024. Association for Computational Linguistics.
\newblock \doi{10.18653/v1/2024.acl-long.714}.
\newblock URL \url{https://aclanthology.org/2024.acl-long.714}.

\bibitem[Fan et~al.(2024)Fan, Ding, Ning, Wang, Li, Yin, Chua, and
  Li]{10.1145/3637528.3671470}
Wenqi Fan, Yujuan Ding, Liangbo Ning, Shijie Wang, Hengyun Li, Dawei Yin,
  Tat-Seng Chua, and Qing Li.
\newblock {A Survey on RAG Meeting LLMs: Towards Retrieval-Augmented Large
  Language Models}.
\newblock In \emph{Proceedings of the 30th ACM SIGKDD Conference on Knowledge
  Discovery and Data Mining}, KDD '24, page 6491–6501, New York, NY, USA,
  2024. Association for Computing Machinery.
\newblock ISBN 9798400704901.
\newblock \doi{10.1145/3637528.3671470}.
\newblock URL \url{https://doi.org/10.1145/3637528.3671470}.

\bibitem[Hagstr{\"o}m et~al.(2025)Hagstr{\"o}m, Marjanovic, Yu, Arora, Lioma,
  Maistro, Atanasova, and Augenstein]{hagstrom-etal-2025-reality}
Lovisa Hagstr{\"o}m, Sara~Vera Marjanovic, Haeun Yu, Arnav Arora, Christina
  Lioma, Maria Maistro, Pepa Atanasova, and Isabelle Augenstein.
\newblock {A Reality Check on Context Utilisation for Retrieval-Augmented
  Generation}.
\newblock In Wanxiang Che, Joyce Nabende, Ekaterina Shutova, and Mohammad~Taher
  Pilehvar, editors, \emph{Proceedings of the 63rd Annual Meeting of the
  Association for Computational Linguistics (Volume 1: Long Papers)}, pages
  19691--19730, Vienna, Austria, July 2025. Association for Computational
  Linguistics.
\newblock ISBN 979-8-89176-251-0.
\newblock \doi{10.18653/v1/2025.acl-long.968}.
\newblock URL \url{https://aclanthology.org/2025.acl-long.968/}.

\bibitem[Hagström et~al.(2025)Hagström, Kim, Yu, Lee, Johansson, Cho, and
  Augenstein]{hagström2025cubbenchmarkingcontextutilisation}
Lovisa Hagström, Youna Kim, Haeun Yu, Sang-Goo Lee, Richard Johansson, Hyunsoo
  Cho, and Isabelle Augenstein.
\newblock {CUB: Benchmarking Context Utilisation Techniques for Language
  Models}, 2025.
\newblock URL \url{https://arxiv.org/abs/2505.16518}.

\bibitem[Hu et~al.(2025)Hu, Cao, Chen, Liu, and
  Zhao]{10.1609/aaai.v39i22.34583}
Chenhui Hu, Pengfei Cao, Yubo Chen, Kang Liu, and Jun Zhao.
\newblock {Knowledge in Superposition: Unveiling the Failures of Lifelong
  Knowledge Editing for Large Language Models}.
\newblock In \emph{Proceedings of the Thirty-Ninth AAAI Conference on
  Artificial Intelligence and Thirty-Seventh Conference on Innovative
  Applications of Artificial Intelligence and Fifteenth Symposium on
  Educational Advances in Artificial Intelligence}, AAAI'25/IAAI'25/EAAI'25.
  AAAI Press, 2025.
\newblock ISBN 978-1-57735-897-8.
\newblock \doi{10.1609/aaai.v39i22.34583}.
\newblock URL \url{https://doi.org/10.1609/aaai.v39i22.34583}.

\bibitem[Islam et~al.(2025)Islam, Atanasova, and
  Augenstein]{islam2025multistepknowledgeinteractionanalysis}
Sekh~Mainul Islam, Pepa Atanasova, and Isabelle Augenstein.
\newblock {Multi-Step Knowledge Interaction Analysis via Rank-2 Subspace
  Disentanglement}, 2025.
\newblock URL \url{https://arxiv.org/abs/2511.01706}.

\bibitem[Kuhn et~al.(2023)Kuhn, Gal, and Farquhar]{kuhn2023semantic}
Lorenz Kuhn, Yarin Gal, and Sebastian Farquhar.
\newblock {Semantic Uncertainty: Linguistic Invariances for Uncertainty
  Estimation in Natural Language Generation}.
\newblock In \emph{The Eleventh International Conference on Learning
  Representations}, 2023.
\newblock URL \url{https://openreview.net/forum?id=VD-AYtP0dve}.

\bibitem[Liao et~al.(2025)Liao, Antoniak, Cheong, Cheng, Lee, Lo, Chang, and
  Zhang]{journals/corr/abs-2411-05025}
Zhehui Liao, Maria Antoniak, Inyoung Cheong, Evie Yu-Yen Cheng, Ai-Heng Lee,
  Kyle Lo, Joseph~Chee Chang, and Amy~X. Zhang.
\newblock {LLMs as Research Tools: A Large Scale Survey of Researchers' Usage
  and Perceptions}.
\newblock In \emph{COLM}, 2025.
\newblock URL \url{https://openreview.net/forum?id=p0BwJk3R1p}.

\bibitem[Marjanovic et~al.(2024)Marjanovic, Yu, Atanasova, Maistro, Lioma, and
  Augenstein]{marjanovic-etal-2024-dynamicqa}
Sara~Vera Marjanovic, Haeun Yu, Pepa Atanasova, Maria Maistro, Christina Lioma,
  and Isabelle Augenstein.
\newblock {{DYNAMICQA}: Tracing Internal Knowledge Conflicts in Language
  Models}.
\newblock In Yaser Al-Onaizan, Mohit Bansal, and Yun-Nung Chen, editors,
  \emph{Findings of the Association for Computational Linguistics: EMNLP 2024},
  pages 14346--14360, Miami, Florida, USA, November 2024. Association for
  Computational Linguistics.
\newblock \doi{10.18653/v1/2024.findings-emnlp.838}.
\newblock URL \url{https://aclanthology.org/2024.findings-emnlp.838/}.

\bibitem[McCoy et~al.(2019)McCoy, Pavlick, and Linzen]{mccoy-etal-2019-right}
R.~Thomas McCoy, Ellie Pavlick, and Tal Linzen.
\newblock {Right for the Wrong Reasons: Diagnosing Syntactic Heuristics in
  Natural Language Inference}.
\newblock In Anna Korhonen, David Traum, and Llu{\'i}s M{\`a}rquez, editors,
  \emph{Proceedings of the 57th Annual Meeting of the Association for
  Computational Linguistics}, pages 3428--3448, Florence, Italy, July 2019.
  Association for Computational Linguistics.
\newblock \doi{10.18653/v1/P19-1334}.
\newblock URL \url{https://aclanthology.org/P19-1334/}.

\bibitem[Mizrahi et~al.(2024)Mizrahi, Kaplan, Malkin, Dror, Shahaf, and
  Stanovsky]{mizrahi-etal-2024-state}
Moran Mizrahi, Guy Kaplan, Dan Malkin, Rotem Dror, Dafna Shahaf, and Gabriel
  Stanovsky.
\newblock {State of What Art? A Call for Multi-Prompt {LLM} Evaluation}.
\newblock \emph{Transactions of the Association for Computational Linguistics},
  12:\penalty0 933--949, 2024.
\newblock \doi{10.1162/tacl_a_00681}.
\newblock URL \url{https://aclanthology.org/2024.tacl-1.52/}.

\bibitem[Ni et~al.(2024)Ni, Bi, Guo, and
  Cheng]{ni2024llmsneedretrievalaugmentation}
Shiyu Ni, Keping Bi, Jiafeng Guo, and Xueqi Cheng.
\newblock {When Do {LLM}s Need Retrieval Augmentation? Mitigating {LLM}s{'}
  Overconfidence Helps Retrieval Augmentation}.
\newblock In Lun-Wei Ku, Andre Martins, and Vivek Srikumar, editors,
  \emph{Findings of the Association for Computational Linguistics ACL 2024},
  pages 11375--11388, Bangkok, Thailand and virtual meeting, August 2024.
  Association for Computational Linguistics.
\newblock \doi{10.18653/v1/2024.findings-acl.675}.
\newblock URL \url{https://aclanthology.org/2024.findings-acl.675}.

\bibitem[Pava et~al.(2025)Pava, Meinhardt, Zaman, Friedman, Truong, Zhang,
  Marivate, and Koyejo]{pava2025mind}
Juan Pava, Caroline Meinhardt, Haifa Badi~Uz Zaman, Toni Friedman, Sang~T
  Truong, Daniel Zhang, Vukosi Marivate, and Sanmi Koyejo.
\newblock {Mind the (Language) Gap: Mapping the Challenges of LLM Development
  in Low-Resource Language Contexts}, 2025.
\newblock URL
  \url{https://hai.stanford.edu/assets/files/hai-taf-pretoria-white-paper-mind-the-language-gap.pdf}.

\bibitem[Petroni et~al.(2019)Petroni, Rockt{\"a}schel, Riedel, Lewis, Bakhtin,
  Wu, and Miller]{petroni-etal-2019-language}
Fabio Petroni, Tim Rockt{\"a}schel, Sebastian Riedel, Patrick Lewis, Anton
  Bakhtin, Yuxiang Wu, and Alexander Miller.
\newblock {Language Models as Knowledge Bases?}
\newblock In Kentaro Inui, Jing Jiang, Vincent Ng, and Xiaojun Wan, editors,
  \emph{Proceedings of the 2019 Conference on Empirical Methods in Natural
  Language Processing and the 9th International Joint Conference on Natural
  Language Processing (EMNLP-IJCNLP)}, pages 2463--2473, Hong Kong, China,
  November 2019. Association for Computational Linguistics.
\newblock \doi{10.18653/v1/D19-1250}.
\newblock URL \url{https://aclanthology.org/D19-1250/}.

\bibitem[Pezeshkpour et~al.(2021)Pezeshkpour, Jain, Wallace, and
  Singh]{pezeshkpour-etal-2021-empirical}
Pouya Pezeshkpour, Sarthak Jain, Byron Wallace, and Sameer Singh.
\newblock {An Empirical Comparison of Instance Attribution Methods for {NLP}}.
\newblock In Kristina Toutanova, Anna Rumshisky, Luke Zettlemoyer, Dilek
  Hakkani-Tur, Iz~Beltagy, Steven Bethard, Ryan Cotterell, Tanmoy Chakraborty,
  and Yichao Zhou, editors, \emph{Proceedings of the 2021 Conference of the
  North American Chapter of the Association for Computational Linguistics:
  Human Language Technologies}, pages 967--975, Online, June 2021. Association
  for Computational Linguistics.
\newblock \doi{10.18653/v1/2021.naacl-main.75}.
\newblock URL \url{https://aclanthology.org/2021.naacl-main.75/}.

\bibitem[Shi et~al.(2025)Shi, Xu, Wang, Qin, Wang, Wang, Wang, Ebrahimi, and
  Wang]{10.1145/3735633}
Haizhou Shi, Zihao Xu, Hengyi Wang, Weiyi Qin, Wenyuan Wang, Yibin Wang, Zifeng
  Wang, Sayna Ebrahimi, and Hao Wang.
\newblock {Continual Learning of Large Language Models: A Comprehensive
  Survey}.
\newblock \emph{ACM Comput. Surv.}, 58\penalty0 (5), November 2025.
\newblock ISSN 0360-0300.
\newblock \doi{10.1145/3735633}.
\newblock URL \url{https://doi.org/10.1145/3735633}.

\bibitem[Shojaee et~al.(2025)Shojaee, Mirzadeh, Alizadeh, Horton, Bengio, and
  Farajtabar]{shojaee2025illusionthinkingunderstandingstrengths}
Parshin Shojaee, Iman Mirzadeh, Keivan Alizadeh, Maxwell Horton, Samy Bengio,
  and Mehrdad Farajtabar.
\newblock {The Illusion of Thinking: Understanding the Strengths and
  Limitations of Reasoning Models via the Lens of Problem Complexity}, 2025.
\newblock URL \url{https://arxiv.org/abs/2506.06941}.

\bibitem[Wang et~al.(2024{\natexlab{a}})Wang, Zhu, Liu, Zheng, Chen, and
  Li]{10.1145/3698590}
Song Wang, Yaochen Zhu, Haochen Liu, Zaiyi Zheng, Chen Chen, and Jundong Li.
\newblock {Knowledge Editing for Large Language Models: A Survey}.
\newblock \emph{ACM Comput. Surv.}, 57\penalty0 (3), November
  2024{\natexlab{a}}.
\newblock ISSN 0360-0300.
\newblock \doi{10.1145/3698590}.
\newblock URL \url{https://doi.org/10.1145/3698590}.

\bibitem[Wang et~al.(2023)Wang, Wei, Schuurmans, Le, Chi, Narang, Chowdhery,
  and Zhou]{conf/iclr/0002WSLCNCZ23}
Xuezhi Wang, Jason Wei, Dale Schuurmans, Quoc~V. Le, Ed~H. Chi, Sharan Narang,
  Aakanksha Chowdhery, and Denny Zhou.
\newblock {Self-Consistency Improves Chain of Thought Reasoning in Language
  Models}.
\newblock In \emph{ICLR}. OpenReview.net, 2023.
\newblock URL \url{https://openreview.net/forum?id=1PL1NIMMrw}.

\bibitem[Wang et~al.(2024{\natexlab{b}})Wang, Gangi~Reddy, Mujahid, Arora,
  Rubashevskii, Geng, Mohammed~Afzal, Pan, Borenstein, Pillai, Augenstein,
  Gurevych, and Nakov]{wang-etal-2024-factcheck}
Yuxia Wang, Revanth Gangi~Reddy, Zain~Muhammad Mujahid, Arnav Arora, Aleksandr
  Rubashevskii, Jiahui Geng, Osama Mohammed~Afzal, Liangming Pan, Nadav
  Borenstein, Aditya Pillai, Isabelle Augenstein, Iryna Gurevych, and Preslav
  Nakov.
\newblock {Factcheck-Bench: Fine-Grained Evaluation Benchmark for Automatic
  Fact-checkers}.
\newblock In Yaser Al-Onaizan, Mohit Bansal, and Yun-Nung Chen, editors,
  \emph{Findings of the Association for Computational Linguistics: EMNLP 2024},
  pages 14199--14230, Miami, Florida, USA, November 2024{\natexlab{b}}.
  Association for Computational Linguistics.
\newblock \doi{10.18653/v1/2024.findings-emnlp.830}.
\newblock URL \url{https://aclanthology.org/2024.findings-emnlp.830/}.

\bibitem[Wei et~al.(2022)Wei, Wang, Schuurmans, Bosma, ichter, Xia, Chi, Le,
  and Zhou]{wei2022chainofthought}
Jason Wei, Xuezhi Wang, Dale Schuurmans, Maarten Bosma, brian ichter, Fei Xia,
  Ed~Chi, Quoc~V Le, and Denny Zhou.
\newblock {Chain-of-Thought Prompting Elicits Reasoning in Large Language
  Models}.
\newblock In S.~Koyejo, S.~Mohamed, A.~Agarwal, D.~Belgrave, K.~Cho, and A.~Oh,
  editors, \emph{Advances in Neural Information Processing Systems}, volume~35,
  pages 24824--24837. Curran Associates, Inc., 2022.
\newblock URL
  \url{https://proceedings.neurips.cc/paper_files/paper/2022/file/9d5609613524ecf4f15af0f7b31abca4-Paper-Conference.pdf}.

\bibitem[Wiegreffe and Pinter(2019)]{wiegreffe-pinter-2019-attention}
Sarah Wiegreffe and Yuval Pinter.
\newblock {Attention is not not Explanation}.
\newblock In Kentaro Inui, Jing Jiang, Vincent Ng, and Xiaojun Wan, editors,
  \emph{Proceedings of the 2019 Conference on Empirical Methods in Natural
  Language Processing and the 9th International Joint Conference on Natural
  Language Processing (EMNLP-IJCNLP)}, pages 11--20, Hong Kong, China, November
  2019. Association for Computational Linguistics.
\newblock \doi{10.18653/v1/D19-1002}.
\newblock URL \url{https://aclanthology.org/D19-1002/}.

\bibitem[Xu et~al.(2025)Xu, Liu, Choi, Smith, and
  Hajishirzi]{xu-etal-2025-infini}
Hao Xu, Jiacheng Liu, Yejin Choi, Noah~A. Smith, and Hannaneh Hajishirzi.
\newblock {Infini-gram mini: Exact n-gram Search at the {I}nternet Scale with
  {FM}-Index}.
\newblock In Christos Christodoulopoulos, Tanmoy Chakraborty, Carolyn Rose, and
  Violet Peng, editors, \emph{Proceedings of the 2025 Conference on Empirical
  Methods in Natural Language Processing}, pages 24955--24980, Suzhou, China,
  November 2025. Association for Computational Linguistics.
\newblock ISBN 979-8-89176-332-6.
\newblock \doi{10.18653/v1/2025.emnlp-main.1268}.
\newblock URL \url{https://aclanthology.org/2025.emnlp-main.1268/}.

\bibitem[Yan et~al.(2025)Yan, Xu, Du, Yao, Wang, Guo, and
  Chen]{journals/corr/abs-2504-00509}
Kai Yan, Yufei Xu, Zhengyin Du, Xuesong Yao, Zheyu Wang, Xiaowen Guo, and
  Jiecao Chen.
\newblock {Recitation over Reasoning: How Cutting-Edge Language Models Can Fail
  on Elementary School-Level Reasoning Problems?}
\newblock \emph{CoRR}, abs/2504.00509, April 2025.
\newblock URL \url{https://arxiv.org/abs/2504.00509}.

\bibitem[Yu et~al.(2024)Yu, Atanasova, and Augenstein]{yu-etal-2024-revealing}
Haeun Yu, Pepa Atanasova, and Isabelle Augenstein.
\newblock {Revealing the Parametric Knowledge of Language Models: A Unified
  Framework for Attribution Methods}.
\newblock In Lun-Wei Ku, Andre Martins, and Vivek Srikumar, editors,
  \emph{Proceedings of the 62nd Annual Meeting of the Association for
  Computational Linguistics (Volume 1: Long Papers)}, pages 8173--8186,
  Bangkok, Thailand, August 2024. Association for Computational Linguistics.
\newblock \doi{10.18653/v1/2024.acl-long.444}.
\newblock URL \url{https://aclanthology.org/2024.acl-long.444/}.

\bibitem[Yu et~al.(2025)Yu, Jeong, Pawar, Shin, Jin, Myung, Oh, and
  Augenstein]{journals/corr/abs-2508-08879}
Haeun Yu, Seogyeong Jeong, Siddhesh Pawar, Jisu Shin, Jiho Jin, Junho Myung,
  Alice Oh, and Isabelle Augenstein.
\newblock {Entangled in Representations: Mechanistic Investigation of Cultural
  Biases in Large Language Models}.
\newblock \emph{CoRR}, abs/2508.08879, August 2025.
\newblock URL
  \url{http://dblp.uni-trier.de/db/journals/corr/corr2508.html#abs-2508-08879}.

\bibitem[Zhao et~al.(2024)Zhao, Ren, Hessel, Cardie, Choi, and
  Deng]{conf/iclr/Zhao0HC0D24}
Wenting Zhao, Xiang Ren, Jack Hessel, Claire Cardie, Yejin Choi, and Yuntian
  Deng.
\newblock {WildChat: 1M ChatGPT Interaction Logs in the Wild}.
\newblock In \emph{ICLR}. OpenReview.net, 2024.
\newblock URL \url{https://openreview.net/forum?id=Bl8u7ZRlbM}.

\end{thebibliography}
\end{document}